\documentclass[10pt,twocolumn,letterpaper]{article}

\usepackage[pagenumbers]{cvpr} 

\usepackage{graphicx}
\usepackage{amsmath}
\usepackage{amssymb}
\usepackage{booktabs}
\usepackage{multirow}
\usepackage{multicol}
\usepackage{lipsum}  
\usepackage{soul}

\usepackage[pagebackref,breaklinks,colorlinks]{hyperref}

\usepackage[capitalize]{cleveref}
\crefname{section}{Sec.}{Secs.}
\Crefname{section}{Section}{Sections}
\Crefname{table}{Table}{Tables}
\crefname{table}{Tab.}{Tabs.}

\def\cvprPaperID{****} 
\def\confName{CVPR}
\def\confYear{2022}

\begin{document}

\title{MegLoc: A Robust and Accurate Visual Localization Pipeline
}

\author{Shuxue Peng\footnotemark[1] , Zihang He\footnotemark[1] , Haotian Zhang\footnotemark[1] , Ran Yan\footnotemark[1] , Chuting Wang\footnotemark[1] , Qingtian Zhu\footnotemark[1] , Xiao Liu\\
Megvii 3D
}

\maketitle
\renewcommand{\thefootnote}{\fnsymbol{footnote}}
\footnotetext[1]{Equally contributed, order determined by a team lottery.}

\begin{abstract}

In this paper, we present a visual localization pipeline, namely \textbf{MegLoc}, for robust and accurate 6-DoF pose estimation under varying scenarios, including indoor and outdoor scenes, different time across a day, different seasons across a year, and even across years. \textbf{MegLoc} achieves state-of-the-art results on a range of challenging datasets, including winning the Outdoor and Indoor Visual Localization Challenge of ICCV 2021 Workshop on Long-term Visual Localization under Changing Conditions, as well as the Re-localization Challenge for Autonomous Driving of ICCV 2021 Workshop on Map-based Localization for Autonomous Driving.

\end{abstract}

\section{Introduction}
Visual localization is a key technology for applications such as Augmented, Mixed, and Virtual Reality, as well as for robotics. It aims to estimate 6-DoF camera poses for query images by aid of a series of mapping images with given ground truth poses. Mapping images and query images suffer large appearance variations caused by seasonal and illuminational changes.

\begin{figure*}[ht]
    \centering
    \includegraphics[width=\textwidth]{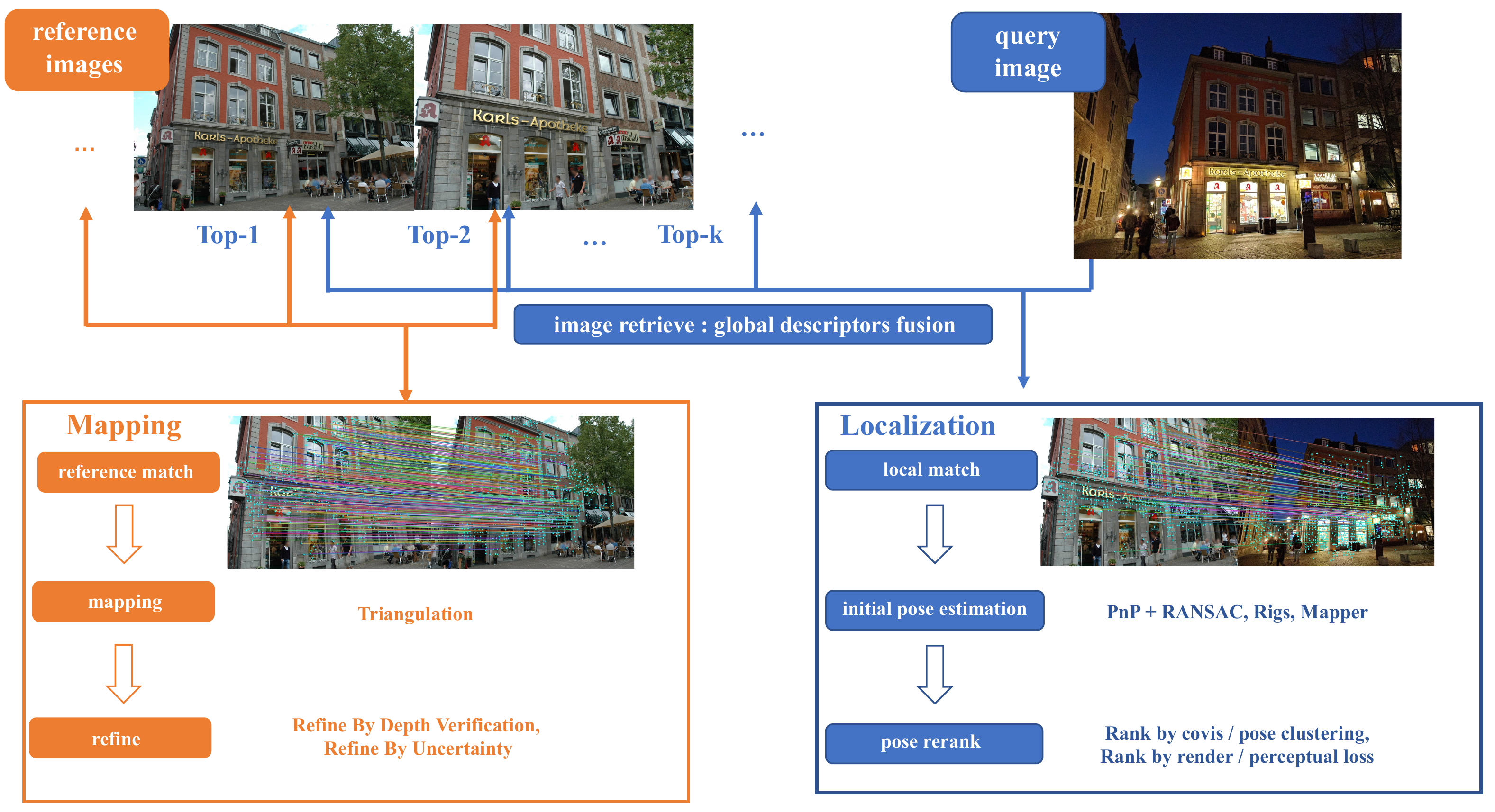}
    \caption{Illustration of the whole pipeline of MegLoc. Specifically, the pipeline is divided into two stages, namely, mapping and localization. For mapping stage, we first do reference matching, then triangulate the matches using COLMAP~\cite{schoenberger2016sfm}, and finally we refine the 3D map by depth verification and uncertainty rejection. For localization stage, we retrieve those reference images similar to the current query image by global descriptors fusion. We perform local feature matching between reference images and the query image. After that we leverage PnP algorithm to compute initial poses and rerank the poses. A final pose optimization step is also performed to obtain the final poses.}
    \label{fig:pipeline}
\end{figure*}

MegLoc follows a conventional two-stage localization fashion, namely mapping and localization. The two stages will get respectively elaborated in the following sections.
MegLoc can function as a complete visual localization pipeline, while also follows a modular design pattern, where each module can be replaced individually. For example, it is possible to use MegLoc's localization module together with another mapping method.
The overall pipeline of MegLoc is illustrated in \cref{fig:pipeline}.

\section{Features}

\subsection{Image Pre-processing}

\paragraph{Resizing}
Empirically, each image is resized to make the larger side 1600 pixels while remaining its original aspect ratio. Some pixels of the bottom-right border might get cropped, to ensure each dimension to be a multiple of 8.

\paragraph{Masking}
Frequently dynamic objects introduce unreliable and unrepeatable keypoints to our solution. To alleviate this problem, we perform semantic segmentation on images and mask out dynamic classes. The selected model is a DeepLab-V3+ network~\cite{chen2018encoder} pretrained on CityScapes dataset~\cite{cordts2016cityscapes}. The definition of dynamic classes is provided by CityScapes officially. For cases that the bottom part of the camera view is always occluded, \eg by the car shell, local features within this area impede subsequent feature matching and triangulation. We remove local features in this area.

\paragraph{Undistortion}

Due to positional encoding, SuperGlue~\cite{sarlin2020superglue} is known to be sensitive to lens distortion. Undistortion of the images before matching is proven by experiments to generate more accurate matches.

\subsection{Local Feature}

SuperPoint~\cite{detone2018superpoint} and ASLFeat~\cite{luo2020aslfeat} are selected as local feature extractors to fully utilize their complimentary characteristics. Specifically, SuperPoint is trained on COCO dataset~\cite{lin2014microsoft}, and it generally works fairly well on different datasets. In contrast, ASLFeat has a more powerful backbone which involves DCN~\cite{dai2017deformable,zhu2019deformable} for shape-awareness local feature extraction. Also, ASLFeat performs multi-level feature fusion.

The Non-Maximum Suppression (NMS) radius and keypoints threshold are tuned empirically based on the experiments conducted in the CVPR 2021 Image Matching Challenge~\cite{bi2021method}. As a result, approximately 1500 keypoints are extracted for each image on average. 

\subsection{Global Feature}

\paragraph{NetVLAD}
Image retrieval has long been a fundamental task in computer vision. Given a query image, a robust image retrieval algorithm is supposed to retrieve the most similar images in the image database, which are called reference images or candidates. This task is challenging because the retrieved results are sensitive to the change of illuminations, shape or even the surroundings under certain circumstances. Efforts have been made to improve the performance.

Traditional algorithms are basically resorting to the local features such as SIFT~\cite{lowe2004distinctive} or ORB~\cite{rublee2011orb}, by clustering algorithms such as K-Means. And then, they embed the current image by Bag of Words or TF-IDF using the clustering result of local features to a specified length of vector.

VLAD is also an excellent traditional image retrieval algorithm. Given an image and its local features of the shape $N\times D$, where $N$ and $D$ respectively denote the number of local features and the length of each local feature vector, VLAD embeds the local feature matrix to a certain representation matrix of shape $K\times D$. $K$ is the predefined number of clustering centroids. Formally, the representation matrix is computed as follows:
\begin{equation*}
V(j, k) = \sum_{i=1}^N a_k(x_i)(x_i(j) - c_k(j)),
\end{equation*}
where $x_{i}(j)$ and $c_{k}(j)$ are the $j$-th dimension of $i$-th descriptor and $k$-th centroid respectively. $a_k(x_i)$ is a binary signal function whose value is 1 if the current descriptor belongs to the centroid or cluster, or 0 otherwise.

NetVLAD~\cite{arandjelovic2016netvlad}, literally, leverages CNNs to obtain a global feature of input image in an end-to-end fashion. The original VLAD breaks the continuities of feature extraction because of the hard assignment of $a_k(x_i)$. To ensure the whole process of feature extraction differentiable, NetVLAD reformulates the extracting equation in original VLAD as follows:
\begin{equation*}
    \overline{a_k}(x_i) = \frac{e^{- \alpha \Vert x_i - c_k \Vert^{2}}}{\sum_{k'} {{ e^{- \alpha \Vert{ x_i - c_{k{'}} }\Vert^2}}}}.
\end{equation*}
The equation above assigns the weight of descriptor $x_i$ to cluster $c_k$ according to the proximity. The more the descriptor approximates the centroids, the larger the value $\overline{a_k}(x_i)$ will be. Apparently, $ \overline{a_k}(x_i)$ ranges between $0$ and $1$, so the function provides a soft way to obtain the global information and thus can be trained end-to-end.
Furthermore, one can expand the square in the function, and we can get the final mathematical definition of NetVLAD:
\begin{equation*}
V(j, k) = \sum_{i=1}^N \frac{e^{w^{T}_k + b_k}}{\sum_{k'} e^{w^{T}_{k'} x_i + b_{k'}}}(x_i(j) - c_k(j)),
\end{equation*}
where $w_k$, $b_k$ and $c_k$ are learnable parameters.

\paragraph{Fusion of Multiple Global Features}
According to our experiments, we find that the retrieval results are only partial because of the nature of NetVLAD. For example, in the dataset of RobotCar, NetVLAD tends to retrieve the images that are under the similar illumination condition as the query image, which hinders the final performance. To alleviate this problem, we resort to a fusion result of global features to make global description more robust. We fuse NetVLAD~\cite{arandjelovic2016netvlad}, DELG~\cite{cao2020unifying}, APGeM~\cite{revaud2019learning,tolias2016particular} and OpenIBL~\cite{ge2020self} features as the final global representation.

\paragraph{Reranking of Global Features}
Inspired by DELG~\cite{cao2020unifying} and Patch-NetVLAD~\cite{hausler2021patch}. We use local feature matching to rerank the retrieval results obtained. To be more specific, given a query image, we first retrieve $M$ images by global descriptors. Then, we leverage local features of these images and SuperGlue~\cite{sarlin2020superglue} to conduct local feature matching. We sort the retrieved images by the number of valid correspondence between the retrieved images and the query image. Finally, we select the top $N$ retrieval images as the final retrieval result.

\subsection{Matching}
We retrain SuperGlue~\cite{sarlin2020superglue} together with its official feature extractor SuperPoint~\cite{detone2018superpoint} in an end-to-end manner on MegaDepth dataset~\cite{li2018megadepth}. More specifically, we split the original SuperPoint~\cite{detone2018superpoint} into two sub-networks. The first one is frozen with the official weights to extract keypoints from images, while the other one is fine-tuned to provide feature descriptions. However, we find this adjustment only advances the model performance slightly, since SuperGlue~\cite{sarlin2020superglue} can already match the given points pretty well.

\subsubsection{Guided Pyramid Matching} 
For those corner cases that the number of matches found by SuperGlue~\cite{sarlin2020superglue} is less than 100, a pyramid extraction strategy is then applied, \eg multiple scales and/or multiple orientations and we might combine the matches in different scales (\textit{ALL}) or trust the one with the most number of matches (\textit{MAX}).

\section{Mapping}
This section introduces MegLoc's mapping strategy under different circumstances.

\subsection{Sparse Reconstruction}
For circumstances without given SfM models, we perform sparse reconstruction on given mapping images and utilize the reconstructed SfM model for localization. The reconstruction consists of 4 stages, namely retrieval of image pairs, local feature matching, triangulation of 2D keypoints and final map refinement.

\subsubsection{Image Pairs Retrieval}
In order to triangulate 2D points into 3D space, we need to acquire image pairs with shared observations. The retrieval strategies vary according to different scenarios.
\paragraph{By Poses}
Since mapping images are provided with ground truth poses, we can assume that spatially adjacent poses are more likely to have shared observations.
\paragraph{By Co-visibility}
The most naive strategy is to retrieve image pairs by making use of image-level co-visibility information.
\paragraph{By Global Descriptors}\label{para:global}
The retrieval of image pairs can naturally fit into a typical image retrieval pipeline, where images are retrieved by ranking pairwise cosine similarity of global descriptors. NetVLAD~\cite{arandjelovic2016netvlad} is chosen as the global descriptor. We also find that fusing multiple global features together in a similar way as \cite{humenberger2020robust} is helpful.

\paragraph{By Temporal Sequence}
For datasets delivered in a temporally-sequential order, we can apparently make the assumption that temporally neighboring images tend to share observations.
\begin{figure*}[ht]
\includegraphics[width=\linewidth]{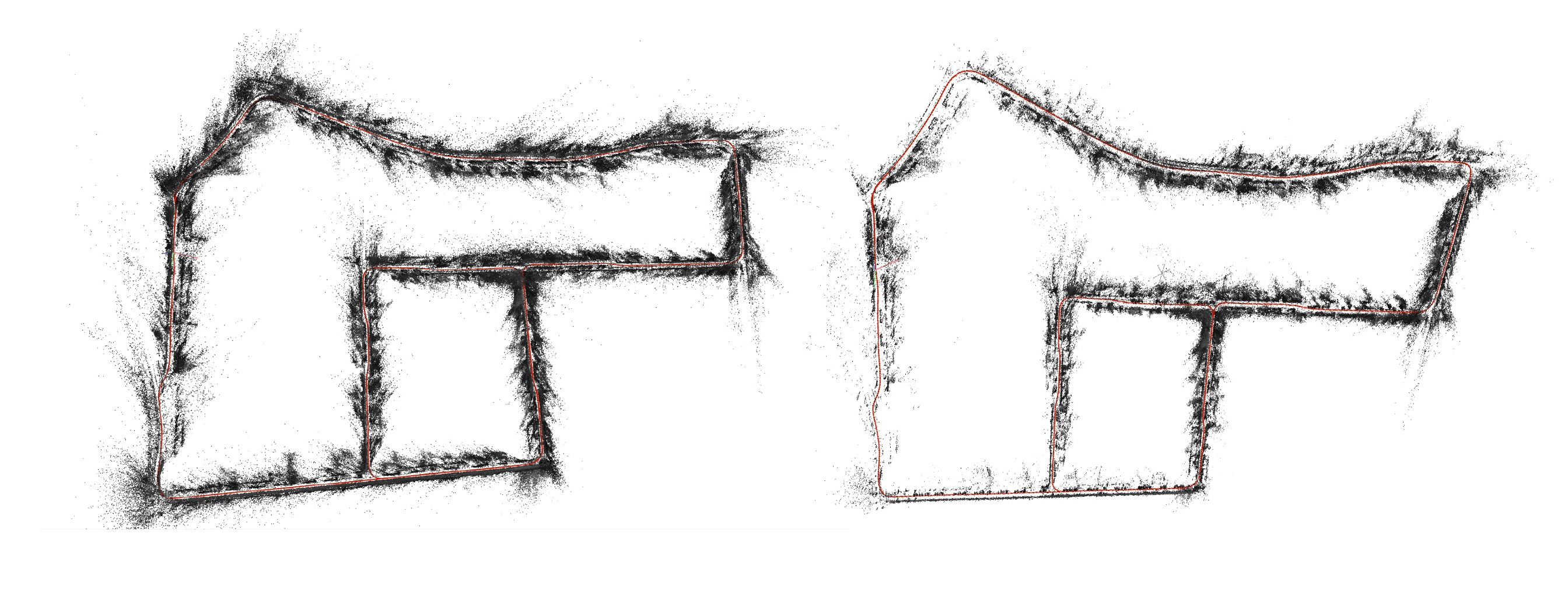}
\caption{An example of 4Seasons dataset~\cite{wenzel2020fourseasons}. Left: original reconstruction of map. Right: reconstructed map with outliers rejection by removing large uncertainty map points.}
\label{fig:1}
\end{figure*}
\subsubsection{3D Map Point Uncertainty \& Map Refinement}

A 3D map point $p_{w}$ in the world frame can be observed in multiple cameras $C_{i} (i=1...N)$ from different views. The camera pose $C_{i}$ and its intrinsic parameters can be represented as $[R_{wc_{i}}, t_{wc_{i}}]$ and $[f_{x}, f_{y}, c_{x}, c_{y}]$, respectively. We assume that a 3D map point $P_{c} = [x, y, z]$ with respect to the camera reference frame corresponds to a observation in the image plane of camera $C_{i}$, and its 2D location can be represented as $p_{uv}$.

The Jacobian of 2D observation $d p_{uv}$ to the 3D map point $dp_{w}$ is 
\begin{equation}
\begin{aligned}
    J_{i} &=\frac{d p_{uv} }{dp_{w}} = \frac{\frac{d p_{uv}}{dp_{c}} \cdot dp_{c} }{dp_{w}}\\
    & = \begin{bmatrix}
\frac{f_{x}}{z} & 0  &  - \frac{x\cdot f_{x}}{z^{2}}\\ 
 0 & \frac{f_{y}}{z}  &  - \frac{y\cdot f_{y}}{z^{2}}
 \end{bmatrix} \cdot R_{wc_{i}}.
\end{aligned}
\end{equation}
Further assuming the observation uncertainty of each pixel on the image plane is an identity matrix, $\Sigma_{uv}=\begin{bmatrix}
1 & 0\\ 
0 & 1
\end{bmatrix}
$, then the information matrix of 3D point $p_{w}$ is 
\begin{equation}
 \Sigma^{-1}_{w_{i}}= J^{T}_{i} \ast \Sigma^{-1}_{uv} \ast J_{i}. 
\end{equation}
The total uncertainty is an addition of the uncertainty of all observations, defined as
\begin{equation}
\Sigma^{-1}_{w} = \Sigma^{-1}_{w_{1}} + \Sigma^{-1}_{w_{2}} + \ldots
\end{equation}

The uncertainty of three orthogonal directions can be obtained by eigen-decomposition, and those map points with large uncertainty can be removed from the map by setting a threshold. The value of the threshold should be determined based on many factors, such as the scale of the reconstructed scene, the number of mapping images, resolution of images. \cref{fig:1} shows the effect of map refinement.

\section{Localization}

\subsection{Image Pairs Retrieval}
The retrieval of image pairs at localization stage is identical to \cref{para:global}, but without ground truth poses.

\subsection{Cluster-wise Localization}

In this section, we illustrate our cluster-wise camera localization strategy.

\subsubsection{Camera Clustering and Reranking}
To resolve the issues caused by repetitive patterns, we purposed two ways of database image clustering, by ground truth 3D poses or by their 3D points co-visibility.

\begin{figure}[ht]
\centering
\includegraphics[width=\linewidth]{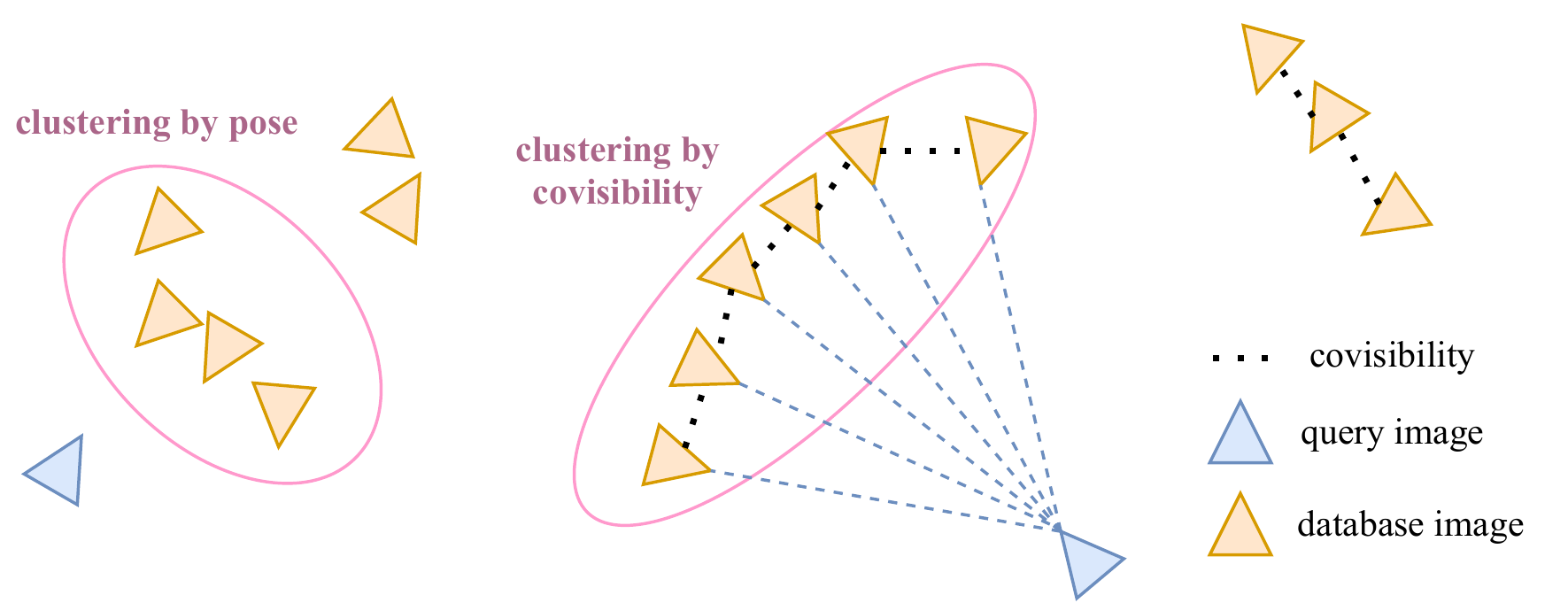}
\caption{Candidate images clustering by pose (left) by 3D points co-visibility (right).}
\label{fig:clustering}
\end{figure}

For clustering by pose, the database images are clustered simply based on their spatial proximity. We only consider the positions while ignore the orientation, since two database images with no overlapping viewing area might still share different co-visible areas with the query image respectively. For clustering by co-visibility, if the number of 3D points co-visible by two database images above a certain threshold, those two images are clustered. This strategy is illustrated in \cref{fig:clustering}.

Many clusters may occur and we estimate the query image pose for each, rather than leveraging all 2D-3D matches into the calculation of PnP.

After that, all potential poses get reranked according to certain criteria, such as number of inliers, and only those database images within the top-1 cluster are kept for final pose refinement. Here we present a specific strategy of camera reranking used under indoor scenes.

In image retrieval, image-wise similarity is measured by global features. These descriptors may suffer from weak texture and repeated pattern and get no longer distinguishable especially under indoor scenes. In these circumstances, the reliability of image-retrieval-based neighborhood searching strategy is seriously challenged. To alleviate this problem, we first align views to be reranked by warping them to the estimated camera pose. In this way, each candidate cluster should have an aligned view to the query image. Then we leverage perceptual similarity~\cite{zhang2018unreasonable} which measures L2 distances between deep feature maps and tends to measure structural similarity. The distance is defined as
\begin{equation}
    d(x,x_0)=\sum_l \frac{1}{H_lW_l}\sum_{h,w}||w_l\odot (\hat{y}^l_{hw}-\hat{y}^l_{0hw})||^2_2,
\end{equation}
where $\hat{y}^l$ denotes the output of a given CNN encoder at layer $l$ and $w$ is a vector for channel-wise scaling. We perform this metric between each one of the aligned candidate views and the query image, and the cluster, whose estimated candidate view is assigned with the lowest score, is selected as the neighborhood for following refinement.
\begin{figure*}[t]
\includegraphics[width=\linewidth]{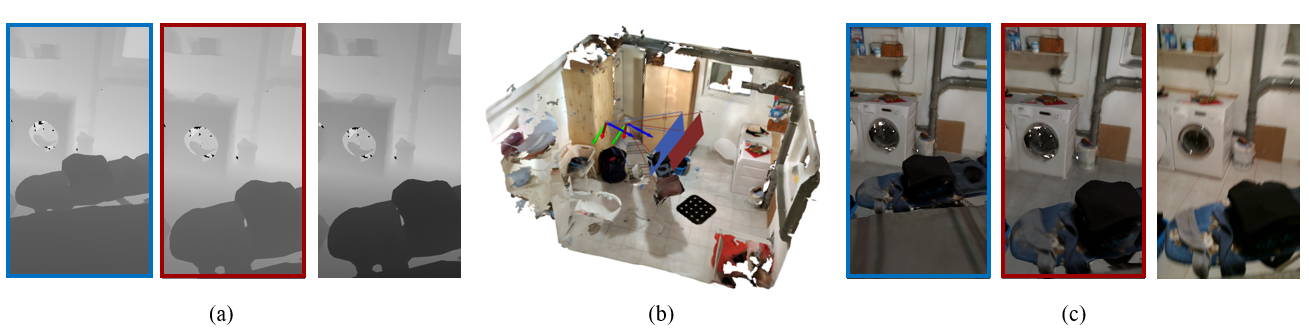}
\caption{Illustration of our ICP-based and rendering-based pose refinement strategy. (a) Rendered depths by the initial camera pose (framed with \textcolor[RGB]{61,89,170}{blue}), by the ICP-based refined pose (framed with \textcolor[RGB]{127,49,41}{red}), and the given reference depth (frameless).
(b) Camera poses observed under the 3D scene.
(c) Rendered RGB images by the initial camera pose (framed with \textcolor[RGB]{61,89,170}{blue}), by the rendering-based refined pose (framed with \textcolor[RGB]{127,49,41}{red}), and the given reference image (frameless).}
\label{fig:2}
\end{figure*}

\subsection{Pose Refinement}
Localization provides preliminary camera poses, which requires further refinement to get more accurate pose. The strategies applied depend on different datasets.

\subsubsection{Iterative Closest Point (ICP)}
Given dense depth maps of query image, we can refine the initial camera pose using ICP (Interactive Closest Point). The fundamental goal of ICP is to estimate an optimal transformation $R$ and $t$ from a point cloud to another. By using ICP, we can optimize those preliminarily estimated camera poses towards the ground truth geometry.

\cref{fig:2} (a) illustrates the camera pose refinement by ICP.

\subsubsection{Refinement by rendering}
We adopt differentiable rendering to optimize the camera pose, whose basic principle is similar to ICP, but the optimization objective is a measurement of photometric similarity instead.
\cref{fig:2} (c) illustrates the camera pose refinement by rendering.

\section{Other Attempts}
Apart from attempts made with a conventional visual localization pipeline, we also make other attempts that take advantage of recent studies in the field of dense reconstruction and neural rendering. 

\subsection{Multi-view Stereo (MVS)}

Multi-view Stereo (MVS) aims to obtain a dense representation of an object or a scene given a series of images. A denser representation is able to provide more anchor points and make localization more robust.

For datasets not providing per-view depth maps, we attempt to estimate a dense depth map for each mapping image to densify the map. Given the camera intrinsics and extrinsics of mapping images, the problem is formulated as a typical multi-view stereo dense reconstruction. 

\subsection{NeRF-W}
Though traditional techniques such as Structure-from-Motion (SfM) have achieved appealing success in 2D-3D mapping. The field of 3D reconstruction has recently great progress in neural rendering. NeRF~\cite{mildenhall2020nerf} models an object or a scene implicitly by fitting a function from location and direction to color and density, and then applies volume rendering to obtain the novel view.  

So we also attempt to represent the map implicitly with the power of NeRF~\cite{mildenhall2020nerf}, with which we are capable of rendering realistic synthetic images at arbitrary views. NeRF-W~\cite{martin2021nerf} is proposed to synthesize novel views with unconstrained images. It divided dynamic scene into two parts. One is ``static" and the other is ``transient". NeRF-W models the two components separately, making it be robust to transient objects. For task of long-term visual localization, we supposed that NeRF-W is able to handle more complicated situations.

\section{Conclusion}
In this paper, we present MegLoc, a robust and accurate 6-DoF pose estimation pipeline.
We introduce the strategies applied by us for feature extraction, feature matching, map refinement, image retrieval, localization and pose refinement in detail.
MegLoc has achieved state-of-the-art performance for many visual localization benchmarks.

{\small
\bibliographystyle{ieee_fullname}
\bibliography{egbib}

\begin{thebibliography}{10}\itemsep=-1pt

\bibitem{arandjelovic2016netvlad}
Relja Arandjelovic, Petr Gronat, Akihiko Torii, Tomas Pajdla, and Josef Sivic.
\newblock Netvlad: Cnn architecture for weakly supervised place recognition.
\newblock In {\em Proceedings of the IEEE conference on computer vision and
  pattern recognition}, pages 5297--5307, 2016.

\bibitem{bi2021method}
Xiaopeng Bi, Yu Chen, Xinyang Liu, Dehao Zhang, Ran Yan, Zheng Chai, Haotian
  Zhang, and Xiao Liu.
\newblock Method towards cvpr 2021 image matching challenge.
\newblock {\em arXiv preprint arXiv:2108.04453}, 2021.

\bibitem{cao2020unifying}
Bingyi Cao, Andre Araujo, and Jack Sim.
\newblock Unifying deep local and global features for image search.
\newblock In {\em European Conference on Computer Vision}, pages 726--743.
  Springer, 2020.

\bibitem{chen2018encoder}
Liang-Chieh Chen, Yukun Zhu, George Papandreou, Florian Schroff, and Hartwig
  Adam.
\newblock Encoder-decoder with atrous separable convolution for semantic image
  segmentation.
\newblock In {\em Proceedings of the European conference on computer vision
  (ECCV)}, pages 801--818, 2018.

\bibitem{cordts2016cityscapes}
Marius Cordts, Mohamed Omran, Sebastian Ramos, Timo Rehfeld, Markus Enzweiler,
  Rodrigo Benenson, Uwe Franke, Stefan Roth, and Bernt Schiele.
\newblock The cityscapes dataset for semantic urban scene understanding.
\newblock In {\em Proceedings of the IEEE conference on computer vision and
  pattern recognition}, pages 3213--3223, 2016.

\bibitem{dai2017deformable}
Jifeng Dai, Haozhi Qi, Yuwen Xiong, Yi Li, Guodong Zhang, Han Hu, and Yichen
  Wei.
\newblock Deformable convolutional networks.
\newblock In {\em Proceedings of the IEEE international conference on computer
  vision}, pages 764--773, 2017.

\bibitem{detone2018superpoint}
Daniel DeTone, Tomasz Malisiewicz, and Andrew Rabinovich.
\newblock Superpoint: Self-supervised interest point detection and description.
\newblock In {\em Proceedings of the IEEE conference on computer vision and
  pattern recognition workshops}, pages 224--236, 2018.

\bibitem{ge2020self}
Yixiao Ge, Haibo Wang, Feng Zhu, Rui Zhao, and Hongsheng Li.
\newblock Self-supervising fine-grained region similarities for large-scale
  image localization.
\newblock In {\em European Conference on Computer Vision}, pages 369--386.
  Springer, 2020.

\bibitem{hausler2021patch}
Stephen Hausler, Sourav Garg, Ming Xu, Michael Milford, and Tobias Fischer.
\newblock Patch-netvlad: Multi-scale fusion of locally-global descriptors for
  place recognition.
\newblock In {\em Proceedings of the IEEE/CVF Conference on Computer Vision and
  Pattern Recognition}, pages 14141--14152, 2021.

\bibitem{humenberger2020robust}
Martin Humenberger, Yohann Cabon, Nicolas Guerin, Julien Morat, J{\'e}r{\^o}me
  Revaud, Philippe Rerole, No{\'e} Pion, Cesar de Souza, Vincent Leroy, and
  Gabriela Csurka.
\newblock Robust image retrieval-based visual localization using kapture.
\newblock {\em arXiv preprint arXiv:2007.13867}, 2020.

\bibitem{li2018megadepth}
Zhengqi Li and Noah Snavely.
\newblock Megadepth: Learning single-view depth prediction from internet
  photos.
\newblock In {\em Proceedings of the IEEE Conference on Computer Vision and
  Pattern Recognition}, pages 2041--2050, 2018.

\bibitem{lin2014microsoft}
Tsung-Yi Lin, Michael Maire, Serge Belongie, James Hays, Pietro Perona, Deva
  Ramanan, Piotr Doll{\'a}r, and C~Lawrence Zitnick.
\newblock Microsoft coco: Common objects in context.
\newblock In {\em European conference on computer vision}, pages 740--755.
  Springer, 2014.

\bibitem{lowe2004distinctive}
David~G Lowe.
\newblock Distinctive image features from scale-invariant keypoints.
\newblock {\em International journal of computer vision}, 60(2):91--110, 2004.

\bibitem{luo2020aslfeat}
Zixin Luo, Lei Zhou, Xuyang Bai, Hongkai Chen, Jiahui Zhang, Yao Yao, Shiwei
  Li, Tian Fang, and Long Quan.
\newblock Aslfeat: Learning local features of accurate shape and localization.
\newblock In {\em Proceedings of the IEEE/CVF conference on computer vision and
  pattern recognition}, pages 6589--6598, 2020.

\bibitem{martin2021nerf}
Ricardo Martin-Brualla, Noha Radwan, Mehdi~SM Sajjadi, Jonathan~T Barron,
  Alexey Dosovitskiy, and Daniel Duckworth.
\newblock Nerf in the wild: Neural radiance fields for unconstrained photo
  collections.
\newblock In {\em Proceedings of the IEEE/CVF Conference on Computer Vision and
  Pattern Recognition}, pages 7210--7219, 2021.

\bibitem{mildenhall2020nerf}
Ben Mildenhall, Pratul~P Srinivasan, Matthew Tancik, Jonathan~T Barron, Ravi
  Ramamoorthi, and Ren Ng.
\newblock Nerf: Representing scenes as neural radiance fields for view
  synthesis.
\newblock In {\em European conference on computer vision}, pages 405--421.
  Springer, 2020.

\bibitem{revaud2019learning}
Jerome Revaud, Jon Almaz{\'a}n, Rafael~S Rezende, and Cesar Roberto~de Souza.
\newblock Learning with average precision: Training image retrieval with a
  listwise loss.
\newblock In {\em Proceedings of the IEEE/CVF International Conference on
  Computer Vision}, pages 5107--5116, 2019.

\bibitem{rublee2011orb}
Ethan Rublee, Vincent Rabaud, Kurt Konolige, and Gary Bradski.
\newblock Orb: An efficient alternative to sift or surf.
\newblock In {\em 2011 International conference on computer vision}, pages
  2564--2571. IEEE, 2011.

\bibitem{sarlin2020superglue}
Paul-Edouard Sarlin, Daniel DeTone, Tomasz Malisiewicz, and Andrew Rabinovich.
\newblock Superglue: Learning feature matching with graph neural networks.
\newblock In {\em Proceedings of the IEEE/CVF conference on computer vision and
  pattern recognition}, pages 4938--4947, 2020.

\bibitem{schoenberger2016sfm}
Johannes~Lutz Sch\"{o}nberger and Jan-Michael Frahm.
\newblock Structure-from-motion revisited.
\newblock In {\em Conference on Computer Vision and Pattern Recognition
  (CVPR)}, 2016.

\bibitem{tolias2016particular}
Giorgos Tolias, Ronan Sicre, and Herv{\'e} J{\'e}gou.
\newblock Particular object retrieval with integral max-pooling of cnn
  activations.
\newblock In {\em International Conference on Learning Representations (ICLR)},
  pages 1--12, 2016.

\bibitem{wenzel2020fourseasons}
P. Wenzel, R. Wang, N. Yang, Q. Cheng, Q. Khan, L. von Stumberg, N. Zeller, and
  D. Cremers.
\newblock {4Seasons}: A cross-season dataset for multi-weather {SLAM} in
  autonomous driving.
\newblock In {\em Proceedings of the German Conference on Pattern Recognition
  ({GCPR})}, 2020.

\bibitem{zhang2018unreasonable}
Richard Zhang, Phillip Isola, Alexei~A Efros, Eli Shechtman, and Oliver Wang.
\newblock The unreasonable effectiveness of deep features as a perceptual
  metric.
\newblock In {\em Proceedings of the IEEE conference on computer vision and
  pattern recognition}, pages 586--595, 2018.

\bibitem{zhu2019deformable}
Xizhou Zhu, Han Hu, Stephen Lin, and Jifeng Dai.
\newblock Deformable convnets v2: More deformable, better results.
\newblock In {\em Proceedings of the IEEE/CVF Conference on Computer Vision and
  Pattern Recognition}, pages 9308--9316, 2019.

\end{thebibliography}
}

\end{document}